# Semantic/Statistical Coupling For Dynamically Enriching Web-based Ontologies

Mohammed Maree
School of Information Technology
Monash University

Mohammed Belkhatir
Faculty of Computer Science
University of Lyon, France

*Abstract*— With the development of the Semantic Web technology, the use of ontologies to store and retrieve information covering several domains has increased. However, very few ontologies are able to cope with the ever-growing need of frequently updated semantic information or specific user requirements in specialized domains. As a result, a critical issue is related to the unavailability of relational information between concepts, also coined *missing background knowledge*. One solution to address this issue relies on the manual enrichment of ontologies by domain experts which is however a time consuming and costly process, hence the need for dynamic ontology enrichment. In this paper we present an automatic coupled statistical/semantic framework for dynamically enriching large-scale generic ontologies from the World Wide Web. Using the massive amount of information encoded in texts on the Web as a corpus, missing background knowledge can therefore be discovered through a combination of semantic relatedness measures and pattern acquisition techniques and subsequently exploited. The benefits of our approach are: (*i*) proposing the dynamic enrichment of large-scale generic ontologies with missing background knowledge, and thus, enabling the reuse of such knowledge in future. (*ii*) dealing with the issue of costly ontological manual enrichment by domain experts. Experimental results in a precision-based evaluation setting demonstrate the effectiveness of the proposed techniques.

*Keywords- large-scale ontologies, dynamic ontology enrichment, missing background knowledge, semantic relatedness, precision-based evaluation*

## 1. INTRODUCTION

The Semantic Web relies heavily on formal ontologies to structure data for comprehensive and transportable machine understanding. Thus, the proliferation of ontologies factors largely in the Semantic Web's success [1]. Generally, ontologies need to be expanded to include new concepts, instances, and relations. This is particularly important when the knowledge captured by the ontology is out-of-date or when it is unable to satisfy the user requirements in specific domains. However, the development of complete and reliable ontologies that can be used in various Semantic Web applications remains a serious problem. Recently, there have been attempts to build general-purpose ontologies that provide wide coverage of generic domains. Examples of these ontologies are (YAGO2[1], Cyc[2], SUMO[3]). Nevertheless[4], the coverage of these ontologies remains far from complete, specifically when dealing with deep domain knowledge. In addition, they require constant maintenance and update to capture up-to-date knowledge about different domains.

To overcome these drawbacks, we propose an automatic statistical/semantic framework to enrich generic ontologies with missing background knowledge. Our source of ontology enrichment is the massive amount of information encoded in texts on the web. First, we exploit statistical techniques based on the Normalized Term Relatedness function to measure the semantic relatedness between terms that are defined in the ontology (i.e. ontology concepts) and other terms that are missing, i.e. not defined in it. Then, based on the obtained semantic relatedness measures, a semantic-based ontology enrichment technique is employed to enrich the ontology according to a list of predefined lexico-syntactic patterns.

The main contributions of our approach are:

- Combining semantic relatedness measures and pattern acquisition techniques for information extraction.
- Automatic enrichment of large-scale generic ontologies from the WWW.
- Dealing with the issue of time-consuming and costly ontological manual enrichment by domain experts.

The rest of the paper is organized as follows. Section 2 presents an overview of previous research in the ontology enrichment area. A general overview of the proposed framework is given in Section 3. Section 4 describes the steps of the proposed framework. Section 5 discusses the precision-based evaluation of our techniques on the task of enriching a large-scale Web ontology (the WorNet ontology). The final section presents the conclusions and outlines the future works.

---

[1] http://www.mpi-inf.mpg.de/yago-naga/yago/
[2] http://www.cyc.com/
[3] http://www.ontologyportal.or

## 2. RELATED WORKS

The purpose of this section is to discuss 'state-of-the-art' ontology enrichment and population systems/approaches. Following [2], we classify existing systems/approaches that address ontology population, as well as ontology enrichment according to the following categorisation criteria:

- **Elements Learned**: some of the studied systems are mainly concerned with learning ontological elements or entities such as concepts, instances, and semantic relations, while others focus only on learning lexical elements such as words. There are also some systems that learn meta-knowledge about how to extract ontological knowledge from input.
- **Input Requirements**. this criterion is concerned with the types of inputs required by the ontology learning systems in order to help knowledge acquisition. Examples of such inputs are: structured data (such as existing ontologies, database schemas, or knowledge bases), semi-structured data (such as dictionaries and XML documents), and unstructured data (such as natural language texts or Web texts).
- **Learning Method**: the approach an ontology learning system adopts in order to extract knowledge is also of particular interest. In general the leaning method consists of the following:
    a. Learning category (supervised vs. unsupervised, online vs. offline)
    b. Learning approach (statistical vs. symbolic, logical, linguistic-based, pattern matching, template-driven and hybrid methods.
    c. Learning task (classification, clustering, rule-learning, concept formation, ontology population)
    d. degree of automation (manual, semi-automatic, cooperative, fully automatic)
    e. Type and amount of user intervention.
- **Type of Output**: another interesting criterion is the type of output that each system produces. Basically, the output of most of the ontology learning approaches and systems is an updated structure or an extended version of the input ontology or knowledge base.
- **Testing & Evaluation Methods**: it is important to evaluate the resulted ontology by the ontology learning system to see whether the newly obtained concepts are located in the correct positions in the hierarchy of the ontology or not. Additionally, evaluating the enrichment and population methods is great interest because other systems may adopt and use methods that prove to be effective and efficient.

For the rest of this section, we will give an overview of the characteristics of several ontology learning and enrichment systems and classify them according to the introduced set of evaluation criteria.

### A. *Text-To-Onto [3]*

Is an ontology learning environment for learning non-taxonomic conceptual relations from text embedded in a general architecture for semi-automatic acquisition of ontologies. It supports as well the acquisition of conceptual structures such as mapping linguistic resources to the acquired structures. The proposed approach uses shallow text processing methods to identify linguistically related pairs of words. In addition, an algorithm for discovering generalized association rules analyses statistical information about the linguistic output. The purpose of this step is to derive correlations at the conceptual level between the newly obtained concept and the concepts that are already defined in the taxonomy. On the other hand, the discovery algorithm determines support and confidence measures for the relationships between these pairs, as well as for relationships at higher levels of abstraction. It also uses the background knowledge from the taxonomy in order to propose relations at the appropriate level of abstraction. According to the authors, the evaluation showed that though their approach is too weak for fully automatic discovery of non-taxonomic conceptual relations, it is highly adequate to help the ontology engineer with modelling the ontology through proposing conceptual relations. Therefore, much work remains to be done in terms of *i)* how to approach the naming and the categorization of relations into a relation hierarchy and *ii)* how to deal with ontological axioms.

### B. *DODDLE II [4]*

DODDLE II is the extended version of DODDLE Domain Ontology Rapid Development Environment. It acquires both taxonomic and non-taxonomic conceptual relationships, exploiting WordNet and domain-specific texts with the automatic analysis of lexical co-occurrence statistics, based on the idea that a pair of terms with high frequency on co-occurrence statistics can have non-taxonomic conceptual relationships. The taxonomic relationship acquisition module does spell match between the input domain terms and WordNet. The spell match links these terms to WordNet resulting in a hierarchically structured set of all the nodes on the path from these terms to the root of WordNet. The non-taxonomic relationship learning module extracts the pairs of terms that should be related by some relationship from domain-specific texts, analyzing lexical co-occurrence statistics, based on WordSpace that is a multi-dimensional, real-valued vector space where the cosine of the angle between two vectors is a continuous measure of their semantic relatedness. To evaluate the system, some case studies have been done in the field of law. A major issue with this approach is that it relies on WordNet to obtain the taxonomic relations. This is because of the limited domain coverage of WordNet lack of semantic knowledge represented in it. Therefore, many of the obtained relations can be missed because they are not defined in WordNet.

*C. OntoLearn [5]*

This system has been developed with the purpose of improving human productivity in the process that a group of domain experts accomplishes in order to find an agreement on *i)* the identification of the key concepts and relationships in the domain of interest and *ii)* providing an explicit representation of the conceptualization captured in the previous stage. OntoLearn employs a set of text-mining techniques to extract relevant concepts and concept instances from existing documents in a Tourism domain, arrange them in sub-hierarchies, and detect relations among such concepts. The produced sub-hierarchies are placed under the appropriate nodes in WordNet manually by the Ontology Engineer. However, as stated by the authors, structuring terms in sub-trees significantly reduces manual work, because only term heads must be linked to the Ontology. The produced results by the system were evaluated based on judgments made by experts in the domain.

*D. ASIUM [6]*

This system learns cooperatively semantic knowledge from texts syntactically parsed without previous manual processing. This knowledge consists in subcategorization frames of verbs and an ontology of concepts for a specific domain. The inputs of ASIUM result from syntactic parsing of texts; they are subcategorization examples and basic clusters formed by head words that occur with the same verb after the same preposition or with the same syntactical role. ASIUM successively aggregates the clusters to form new concepts in the form of a generality graph that represents the ontology of the domain. Subcategorization frames are learned in parallel, so that as concepts are formed, they fill restrictions of selection in the subcategorization frames. ASIUM proposes a cooperative ML method, which provides the user with a global view of the acquisition task and also with acquisition tools like automatic concept splitting, example generation, and an ontology view with attachments to the verbs.

*E. SYNDIKATE [7]*

This system automatically acquires knowledge from real-world texts and transfers their content to formal representation structures which constitute a corresponding text knowledge base. SYNDIKATE relies on two major kinds of knowledge:

- Grammatical knowledge for syntactic analysis is given as a fully lexicalized dependency grammar. Such a grammar mainly consists of the specification of local valency constraints between lexical items and a possible syntactic modifier. Valency constraints also include restrictions on word order, compatibility of morphosyntactic features, as well as semantic integrity conditions.
- Conceptual knowledge where a domain ontology consists of a set of concept names and a subsumption relations. Concepts are linked by conceptual relations. The corresponding set of relation names denotes conceptual relations which are also organized in a subsumption hierarchy.

The system integrates requirements from the analysis of single sentences, as well as those of referentially linked sentences forming cohesive texts. Besides centring-based discourse analysis mechanisms for pronominal, nominal and bridging anaphora, SYNDIKATE is supplied with a learning module for automatically boot-strapping its domain knowledge as text analysis proceeds. The approach to learning new concepts as a result of text understanding builds on two different sources of evidence: the prior knowledge of the domain the texts are about and grammatical constructions in which unknown lexical items occur in the texts. A given ontology is incrementally updated as new concepts are acquired from real-world texts. The acquisition process is centred on the linguistic and conceptual "quality" of various forms of evidence underlying the generation and refinement of concept hypotheses. On the basis of the quality of evidence, concept hypotheses are ranked according to credibility and the most credible ones are selected for assimilation into the domain knowledge base [2].

*F. Learning Concept Hierarchies from Text Corpora using Formal Concept Analysis [8]*

The proposed approach aims at automatic acquisition of taxonomies or concept hierarchies from a text corpus. It is based on Formal Concept Analysis (FCA), a method mainly used for the analysis of data (deriving implicit relationships between objects described through a set of attributes on the one hand and these attributes on the other). In order to derive context attributes describing the interesting terms, the authors make use of syntactic dependencies between the verbs appearing in the text collection and the heads of the subject, object and PP-complements they subcategorize. The approach was evaluated by comparing the resulting concept hierarchies with hand-crafted taxonomies for two domains: tourism and finance. Besides, it was directly compared with hierarchical agglomerative clustering as well as with Bi-Section-KMeans as an instance of a divisive clustering algorithm. Experimental results demonstrated that the proposed approach outperformed both algorithms and produced better results using the datasets from the two domains.

*G. Automatising the learning of lexical patterns: An application to the enrichment of WordNet by extracting semantic relationships from Wikipedia [9]*

The presented approach is used to identify lexical patterns that represent semantic relationships between concepts in an on-line encyclopedia (Wikipedia). These patterns are then applied to extend an existing ontology (WordNet 1.7) with hyperonymy, hyponymy, holonymy and meronymy relations. The followed procedure consisted of the next steps:
1. Entry sense disambiguation: this step consists in pre-processing the Wikipedia definitions and associating each Wikipedia entry to its corresponding WordNet synset, so the sense of the entry is explicitly determined.

2. Pattern extraction: for each entry, the definition is processed looking for words that are connected with the entry in Wikipedia by means of a hyperlink. If there is a relation in WordNet between the entry and any of those words, the context is analysed and a pattern is extracted for that relation.
3. Pattern generalisation: the patterns extracted in the previous step are compared with each other, and those that are found to be similar are automatically generalised.
4. Identification of new relations: the patterns are applied to discover new relations other than those already present in WordNet.

Experimentally, the precision of relationships depends on the degree of generality chosen for the patterns and the type of relation. Generally, it was around 60–70% for the best combinations proposed.

*H. Using Social Media for Ontology Enrichment [10]*

The proposed approach exploits social media tags, which are crawled from existing social media applications, similarity measures, the DBpedia knowledge base, a disambiguation algorithm and other heuristics to enrich existing ontologies with new semantic information. LT4eL domain ontology on computing that was developed in the Language Technology for eLearning project[4]. It contains 1002 domain concepts, 169 concepts from OntoWordNet and 105 concepts from DOLCE Ultralite. The connection between tags and concepts is established by means of language-specific lexicons, where each lexicon specifies one or more lexicalizations for each concept. A crawler that uses APIs provided by the social networking applications is used to get information about users, resources and tags. The crawler extracts links to resources from social media applications such as Delicious, YouTube and Slideshare together with the tags used to classify the resources and information about the social connections developed inside these web sites. On the other hand, during the enrichment process, similarity measures such as term co-occurrence and cosine similarity are employed to identify tags that are related to the lexicalization of concepts already existing in the LT4eL domain ontology. To identify relations among the existing LT4eL concepts and new concepts derived from the tags, background ontology (DBpedia) is used. Several heuristics are employed to discover taxonomic relations, synonyms and new relations explicitly coded in DBpedia. To evaluate the proposed methodology, three different ontologies (LT4eL computing ontology, a manually enriched ontology which takes the LT4eL one as basis, and an automatically enriched ontology) were used and compared.

*I. RelExt [11]*

RelExt is a system capable of automatically identifying relevant triples (pairs of concepts connected by a relation) over concepts from an existing ontology (from the football domain). It works by extracting relevant verbs and their grammatical arguments (terms) from a domain-specific text collection and computing corresponding relations through a combination of linguistic and statistical processing. For the linguistic annotation, the authors used the SCHUG-system [12], which provides a multi-layered XML format for a given text, specifying dependency structure along with grammatical function assignment, phrase structure, part-of-speech and lemmatization (including decomposition, which is useful in particular for German where compound nouns are often used). For the statistical processing, the authors performed several computations on the extracted data, starting from relevance ranking, and cross-referencing relevant nouns and verbs with the predicate-argument-pairs, to computing co-occurrence-scores in order to construct triples that are specifically used in the football domain. To evaluate the systems, the authors measured its performance against a gold standard that they constructed to benchmark different parameters. Specifically, they divided up the corpus into 4 equally sized sub-corpora of 300 documents, from which we used one sample for benchmark construction.

*J. Ontology Concept Enrichment via Text Mining [13]*

The proposed approach by Wang et al., focuses on enriching the vocabulary of the ontology from domain-relevant documents. For each candidate concept, they measure the similarity between its context and the contexts of the concepts that are defined in the ontology. Then, they use the kNN method to rank the final results to get the top pairs of closest concepts. In order to test the proposed ontology enrichment solution, the authors used a set which contained 530 concept names and 191 test synonyms. Promising results has been obtained compared to existing WordNet-based similarity measures. However, as stated by the authors, additional experiments need to be carried out in order to validate the effectiveness of the proposed solution.

Table 1 shows a classification of the studied ontology learning and enrichment systems/approaches according to the list of classification criteria discussed in section 2.

**Table 1. Classification of the Studied Ontology Learning and Enrichment Systems**

| Index | System/Approach | Elements Learned | Input Requirements | Learning Method | Output | Testing & Evaluation Method |
|---|---|---|---|---|---|---|
| **A** | **Text-To-Onto** | Concepts, Taxonomic and | Web documents, Domain-independent | Formal Concept Analysis, | RDF extensions, Directed | Empirical evaluation and |

---

[4] http://www.lt4el.eu

| | | Non-taxonomic conceptual relations | texts, Semi-structured documents, Structured documents | Clustering | graphs | testing for different domains |
|---|---|---|---|---|---|---|
| B | **DODDLE II** | Taxonomic relations, Non-taxonomic conceptual relations | Domain-specific texts | Analysis of lexical co-occurrence statistics | Hierarchies (sub-trees) and Concept pairs | Evaluated in the Law domain |
| C | **OntoLearn** | Named entities, Sub-trees | Natural Language texts from different domains | Exploits both linguistic and statistical analysis techniques | Sub-hierarchies, Relations between the concepts | Evaluated using manual inspection and human judgments on the results |
| D | **ASIUM** | Verb subcategoy frames, Hierarchies | Unstructured (corpora) | Syntactic analysis and conceptual clustering | Domain-specific Hierarchies | Tested for cooking and patents domains, Terrorism IE |
| E | **SYNDIKATE** | Concept, Taxonomic and non-taxonomic conceptual relations | Unstructured texts | Discourse analysis and Semantic analysis | Knowledge base, Updated input ontology | Evaluated in the I.T. and Medical domains |
| F | **Learning Concept Hierarchies from Text Corpora using Formal Concept Analysis** | Taxonomies and Concept hierarchies | Text corpus | Syntactic and Formal concept analysis | Concept lattice converted into concept hierarchies | Tested using reference ontologies and modelled ontologies by experienced ontology engineer |
| G | **Automatising the learning of lexical patterns: An application to the enrichment of WordNet by extracting semantic relationships from Wikipedia** | Entities and relationships | Natural language texts obtained from Wikipedia encyclopedia | Pattern-based analysis | Entities and relationships between them | Evaluated using hyponymy, hyperonomy, holonomy and meronymy relations |
| H | **Using Social Media for Ontology Enrichment** | Concept, Taxonomic and non-taxonomic conceptual relations | Social media tags obtained from social websites | Statistical analysis, Semantic analysis | Enriched ontology | Tested using three different ontologies |
| I | **RelExt** | Triples in the form of concept-relation-concept | Domain-specific text collection, Predefined ontology | Linguistic and statistical processing | Extended domain ontology | Evaluated using a manually constructed corpus |
| J | **Ontology Concept Enrichment via Text Mining** | Concepts | Domain-related documents | Linguistic and context-based similarity | Enriched ontology | Tested using a set which contained 530 concept names and 191 test synonyms |

As shown in Table 1, the type of input requirements varies from one system to another. Some systems use unstructured domain-independent texts as their input (in some systems this is called: starting point), while others are concerned with texts relevant to the domain of interest. In most of the systems, the learned elements are new concepts and taxonomic and non-taxonomic relations.

Some systems organize the learned elements into sub-trees or concept hierarchies (e.g., OntoLearn and ASIUM). Clearly, statistical analysis and pattern-recognition techniques are the dominant techniques that have been used by most of these systems and approaches. Since the ultimate goal of these systems is to enrich existing ontologies or knowledge bases with additional background semantic knowledge, they all produced such semantic knowledge in the form of concept hierarchies, enriched ontologies, or simply individual concepts and with the semantic and taxonomic relations between them. For the later, the ontology engineer or the domain expert has to manually place these concepts and relations in the correct positions in the source ontology or knowledge base.

Considering the evaluation and testing of the studied systems, we can see that the systems were evaluated and tested using documents or ontologies from specific domains such as Medicine, I.T., and Terrorism domains. To judge the quality of the produced results by each of these systems two different methods were followed: (1) using manual inspection and human judgments, and (2) automatically using precision and recall indicators. Although automatic judgment can be seen to be more effective than the manual approach, it can be argued that human judgments are more accurate and hence may be still required in ontology learning or enrichment approaches.

In the next section, we introduce semantic search and discuss the important role that ontologies play in this field. In addition, we present a detailed discussion on different semantic search approaches, namely those that are dedicated for searching the Web

## 3. GENERAL OVERVIEW

As shown in Figure 1, we propose an automatic statistical/semantic framework to enrich generic ontologies with missing background knowledge. In this figure, we provide the overall organization of the framework which consists of:

- The Named Entity Recognition and Natural Language Processing module where a named entity recognizer is employed to obtain named entities from the text. Next, n-gram term tokenization is utilized where unigrams, bigrams and trigrams are obtained and compared with the ontology concepts. This module is further detailed in section 4-A.
- The Missing Background Knowledge Handler that relies on a statistical technique to measure the semantic relatedness between concepts that are defined in the ontology and terms that are missing from it. This technique is based on the Normalized Term Relatedness and N-gram Hits functions presented in section 4-B.
- The Semantic Ontology Enrichment module aiming at enriching the ontology with axioms based on: (i) the semantic relatedness measures obtained from the previous step, (ii) a predefined list of lexico-syntactic patterns. This technique is discussed in section 4-C.

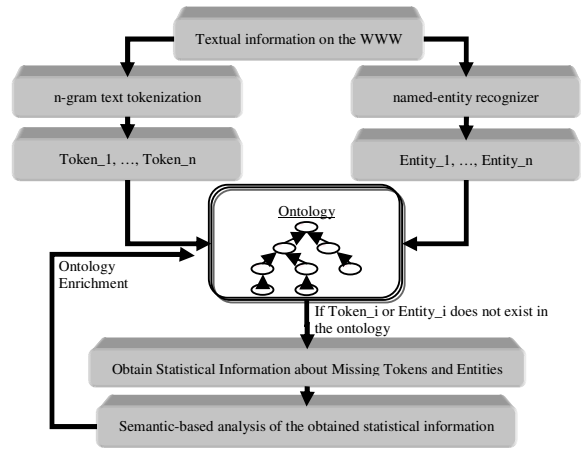

**Fig. 1. Steps of the proposed ontology enrichment framework**

## 4. DETAILED STEPS OF THE PROPOSED FRAMEWORK

Before we detail the steps of the proposed framework, we define what an ontology is, introduce the ontology enrichment process, and present the Normalized Term Relatedness and N-gram Hits functions.

**Definition 1**: An ontology O is a *4-tuple* ⟨C, R, I, A⟩ such that:
- $C=\{(c_i), i \in [1, Card(C)]\}$ represents the set of domain concepts of the ontology.
- $R=\{(r_i), i \in [1, Card(R)]\}$ represents the set of semantic relations holding between the ontology concepts. We consider synonymy, hypernymy, hyponymy, meronymy and holonymy relations.

- I is the set of instances or individuals.
- A is the set of axioms verifying: A={$(r_i,c_j,c_k)$} s.t. i∈[1,Card(R)], j,k∈[1,Card(C)], $c_j,c_k$∈C and $r_i$∈R.

**Definition 2**: Ontology Enrichment: An ontology enrichment algorithm takes a given text corpus $\zeta$ and a given concept hierarchy from O as input and produces for each c ∈ C a set S(c) ⊆ T($\zeta$) where:

- S(c) is the set of suggested enrichment candidates for c. A suggested candidate $t \in T(\zeta)$ is a word or compound word from $\zeta$
- T($\zeta$) is the set of words from $\zeta$

The set of suggested candidates S(c) can be obtained using the NTR function and based on a threshold value $v$ using equation (1).

$$S(c) = \{t \in T(\zeta) \mid NTR(c,t) \geq v\} \quad (1)$$

**Definition 3**: Given a term missing in the ontology $t_{miss}$ and a term in the ontology $t_{in}$ (i.e. an ontology concept), the Normalized Term Relatedness (NTR) function, noted NTR($t_{miss},t_{in}$), is defined as:

$$NTR(t_{miss},t_{in}) = 1 - \frac{\frac{\max\{\log f_1(t_{miss}), \log f_1(t_{in})\} - \log f_2(t_{miss},t_{in})}{\log N_{wd} - \min\{\log f_1(t_{miss}), \log f_1(t_{in})\}}}{\sum_{t_i,t_j} \frac{\max\{\log f_1(t_i), \log f_1(t_j)\} - \log f_2(t_i,t_j)}{\log N_{wd} - \min\{\log f_1(t_i), \log f_1(t_j)\}}}$$

where $f_1$ provides the frequency of a term on the Web, $N_{wd}$ is the total number of indexed web documents considered, $f_2$ is the frequency of co-occurrence of a pair of terms, the $t_i$s are all the considered missing terms and the $t_j$s are the ontology terms.

This measure, modified from its initial formulation as the Google Retrieval Distance [14], considers the number of retrieval results returned by Web search engines (Google, Altavista) to compute a distance between terms. The latter are represented by their text-based descriptions that are used to query the search engines.

This computation is necessary to examine how closely related two terms are by analyzing pairwise co-occurrence frequencies. A distance of one indicates that the two terms always appear together.

**Definition 4**: Given an n-gram, the N-gram Hits (NH) function returns the number of its corresponding search hits where each of the n-gram components are used as search terms in a Web retrieval engine (e.g. Google).

*A. Named Entity Recognition and Natural Language Processing (NLP)*

The first step towards information extraction consists of applying named entity recognition. To do so, we employ GATE, a syntactical pattern matching term recognizer [15]. Although the coverage of GATE is limited to a certain number of named entities, additional rules can be defined manually in order to expand its coverage. However, this process can be tedious and time consuming. Therefore, we exploit the ontology itself as part of the named entity recognition step. In this manner, terms that are not defined in GATE are submitted to the ontology to check whether they are defined in it or not. Next, we apply the following NLP steps on the extracted texts:

1. Stop word removal: some words (e.g. *a*, *an*, *the*, *of*,...) occur frequently in the contextual information without being meaningful for our processing. Therefore, those words are removed based on a predefined stop word list. For more details on this NLP step and how it relates to our approach, please refer to the work presented in [16].
2. N-gram text tokenization: the input text is split into tokens of lengths from 1 to 3. Then, each token is compared against the ontology concepts and instances. Those that do not exist in the ontology are part of the missing background knowledge and considered for further processing by the statistical techniques detailed in section 4-B.
3. Part-of-speech tagging: each token that exists in the ontology is tagged by its grammatical category(ies) (e.g. noun, verb, adjective, etc). For instance, the token *book* belongs to two different grammatical categories: noun and verb.

The following example illustrates the NLP steps mentioned above. In this example we exploit the WordNet ontology as a supplementary source of information that is utilized for named entity recognition.

**Example 1: Article on Java Island taken from Wikipedia**

*Java (Indonesian: Jawa) is an island of Indonesia and the site of its capital city, Jakarta. Once the centre of powerful Hindu-Buddhist kingdoms, Islamic sultanates, and the core of the colonial Dutch East Indies, Java now plays a dominant role in the economic and political life of Indonesia, home to a population of 130 million in 2006.*

First, stop words are removed based on the predefined stop word list. For example, words *the*, *an*, *a* and characters *(*, *)* and *:* are removed from the input text. Then, the n-gram tokenization algorithm tokenizes the input text into corresponding unigrams, bigrams and trigrams. After this step, each n-gram is compared against the ontology concepts to check whether it is already defined in it. The algorithm identifies that the following concepts already exist in WordNet: "Java, Indonesian, Island, Indonesia, site, capital, city, capital city, Jakarta, once, centre, powerful, Hindu, Kingdoms, Islamic, sultanates, core, colonial, Dutch, East, Indies, East Indies, Dutch East Indies, now, plays, dominant, role, economic, political, life, Home, population, 130, million, 2006". The rest of the n-grams are considered as missing background knowledge from the ontology.

*B. Statistical Information about Missing n-grams*

For missing n-grams, a statistical function is utilized to verify whether they can be suggested as candidates to enrich the ontology. To do so, we utilize the NH function. In this context, exact match queries in the form of *Q*= "n-gram" are submitted to the Google retrieval engine and the number of hits for each n-gram is collected. For example, to collect the number of Google hits for the bigram "Hindu-Buddhist Kingdoms", the query Q="Hindu-Buddhist Kingdoms" is submitted to the Google search engine. The output of this step is a set of n-grams each associated to the number of their Google hits. The latter is suggested as the set of candidate inputs for the NTR function. Algorithm 1 demonstrates the process of acquiring the number of Google hits for each n-gram.

---

**Algorithm 1: N-gram Hits Computation**

**Input**: vector of n-gram terms, $V_{n\text{-}gram}$

**Output**: vector of n-grams with number of Google hits > 0, $Results\ V^+_{n\text{-}gram}$

1:  String [] Results
2:  int value
3:  for each n-gram $n_i$ in $V_{n\text{-}gram}$
4:     value=submitQuery("$n_i$")
5:     if(value>0)
6:        $V^+_{n\text{-}gram}$ .add("$n_i$")
7:     end if
8:  end loop

---

For each candidate n-gram, the NTR function measures its semantic relatedness to other terms that were identified by both the named entity recognizer and the ontology. Algorithm 2 details the procedure of acquiring statistical information about missing knowledge in the ontology.

---

**Algorithm 2: Normalized Term Relatedness Computation**

**Input**: set $T_{miss}$ of suggested n-gram terms after applying the NH function, set $T_{in}$ of terms existing in the ontology

**Output**: semantic relatedness measures between terms of $T_{miss}$ and $T_{in}$

1:  int[] result
2:  for each $t_{miss} \in T_{miss}$
3:     for each $t_{in} \in T_{in}$
4:        result[i]= NTR($t_{miss},\ t_{in}$)
5:     end for
6:  end for

---

The NTR function takes as input pairs of terms and produces as output the measures of semantic relatedness between them. For example, Table 1 shows the semantic relatedness measures between the missing term *jawa* and the terms of the WordNet ontology *Java*, *island* and *Indonesia*. It is important to mention that these terms represent a focused subset of the WordNet ontology to which each missing entity is compared. Furthermore, we not only compare the term *jawa* to these terms, but also to the other terms that have been found to be defined in WordNet (section 4-A).

**Table 2. Relatedness measures for the missing term "*jawa*"**

| Wordnet Term / Missing Term | *Java* | *Island* | *Indonesia* |
|---|---|---|---|
| jawa | 0.72 | 0.56 | 0.69 |

*C. Semantic-Based Analysis of Statistical information*

The purpose of this step is to derive semantic information from the results obtained after applying the NTR function. To do so, we define a list of lexico-syntactic patterns that are exploited to find the semantic relation between each missing candidate term and other terms that exist in the ontology (i.e. the ontology concepts). We define patterns to derive synonymy, hypernymy, hyponymy, meronymy and holonymy relations. Examples of the defined patterns are: "X is (the) same as Y", "X is a(n) Y", "X is an instance of Y" and "X is (a) part of Y". These types of relations can be automatically obtained by utilizing the Semantic Relation Extractor (SRE) module. For each pair of semantically related terms, the latter returns the number of hits by submitting each of the patterns to the search engines. Then, relations between pairs of terms are suggested based on the highest values returned.

**Algorithm 3: Semantic Relation Extractor**

**Input**: Semantically related concepts (missing terms are included in $T_{miss}$, terms that exist in the ontology are in $T_{in}$)
**Output**: Suggested relations between terms

1:    String [] suggestedRelations
2:    String [] Patterns
3:    int[] value
4:    for each $t_{miss} \in T_{miss}$
5:      for each $t_{in} \in T_{in}$
6:        for each pattern $p_i$ in Patterns
7:         value.add(submitQueryPattern("$t_{miss}$", $p_i$, "$t_{in}$"))
8:        end for
9:        suggestedRelations.add(max(value))
10:     end for
11:   end for

Algorithm 3 shows the steps of the SRE module. It takes as input pairs of terms with highest semantic relatedness measures and produces as output the suggested semantic relation between them based on the lexico-semantic patterns. For each pattern, the submitQueryPattern function (cf. line 7) submits exact match queries including both terms and returns the number of corresponding search hits. Based on the number of hits returned by the search engine for the queries, relations defined in the patterns with the maximum number of search hits are suggested to be used (cf. line 9). We consider both singular and plural forms of the terms. Furthermore, patterns including negation operators such as "No $t_{miss}$ is a(n) $t_{in}$" were excluded. For example, to extract a hyponymy relation between two terms, we build on the definition of this relation in [17]: a term represented by the synset {x, x', …} is said to be a hyponym of the term represented by the synset {y, y', …} if native speakers of English accept sentences constructed from such frames as "A(n) x is a (kind of) y".

**Example 2: A corporate body is an organization or group of persons that is identified by a particular name, and that acts, or may act, as an entity.[5]**

From Example 2 we find that the term "Corporate Body" is missing from WordNet. However, based on the results of the statistical technique detailed in section 4-B, we are able to assess that this term is related to the concept "Organization" which is already defined in WordNet. To determine the semantic relation holding between these two terms, we utilize the SRE by submitting patterns in the form of exact match queries $Q_i$ such as:

- $Q_1$: "corporate body is an organization", which outputs 80,700 hits
- $Q_2$: "corporate body is a kind of organization", which outputs 0 hits
- $Q_3$: "corporate body is a part of an organization", which outputs 0 hits
- $Q_4$: "corporate body is an instance of an organization", which outputs 0 hits
- $Q_5$: "corporate body is similar to (the same as) an organization", which outputs 0 hits

Based on the number of hits returned by the search engine for the queries $Q_i$, relations defined in the patterns are suggested to be used to enrich the ontology with the term "Corporate Body". The latter is consequently considered as a novel ontology concept (since it is not an instance of the concept "Organization").

When applying this step, we notice that for some of the semantically related terms, no search results were returned for any of the patterns. However, we know from the obtained statistical information that they are semantically related. Therefore, in this case, we define an axiom such that the two terms are related through the "Related To" semantic relation to enrich the ontology.

---

[5].http://www.itsmarc.com/crsbeta/mergedProjects/scmshelf/scmshelf/g_220_corporate_bodies_shelf.htm

For instance, in Example 1 of section 4-A, the algorithm returns that the term "Hindu-Buddhist" is missing from WordNet. Using the NH function, the number of hits returned for this term is 128,000. The NTR function further measures its semantic relatedness to both terms defined according to the named entity recognizer and terms in the ontology. Based on its results, the strongest semantically related term to "Hindu-Buddhist" is "Indonesia". Yet, when applying the SRE, the number of returned results was 0 for all of the defined patterns. Therefore, although none of the used patterns represented the type of semantic relation that holds between these two terms, we are still able to enrich the ontology with the term "Hindu-Buddhist" by relating it to the term "Indonesia" through the "Related To" relation.

Formally, when the missing term is not found to be an instance of an ontology concept through the dedicated pattern, this process generates a set of axioms:

$A_{enrich}=\{(r,c_i,c\_miss_j)\}$ s.t. $r \in R \cup \{\text{"related to"}\}$, $c\_miss_i$ is a missing concept in the knowledge bases (i.e. $c_i \notin C$) and $c_j \in C$.

### D. Generic Ontology Enrichment

Enriching the ontology with new concepts requires finding the appropriate position in the hierarchy of the ontology for locating them, i.e. we need to find the concept in the hierarchy of the ontology concepts that is the most closely related to the new concept. To do this, we compare each new concept with the concepts in the path(s) that originate from the semantically related concept. This comparison enables us to locate the new concept to be added in the proper semantic path in the ontology. Below we summarize the different cases that we consider for the enrichment.

**Case 1:** Concept *C* is missing from the ontology, and C is semantically related to the concept *X* which already exists in the ontology. After applying the semantic-based technique, we find that *C* is related to the concept *X* by one of the semantic relations in the defined patterns. The concept *X* has only one sense in the generic ontology, i.e. there is only one semantic path that originates from *X*. In this case, we update the hierarchy of the semantic path by attaching the missing concept *C* to the concept *X* using the derived semantic relation between them. For example, the concept "Concept" in WordNet has only one sense. Therefore, any term considered as a sub-concept of this concept will be directly located under it. Figure 2 illustrates this case.

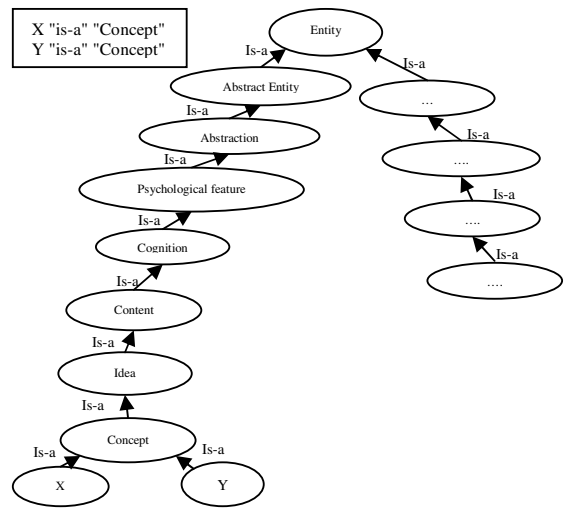

**Fig. 2. Enriching the ontology with missing background knowledge**

**Case 2:** Concept *C* is missing from the ontology, and C is semantically related to the concept X which is already defined in the ontology. After applying the semantic relatedness measure technique, we find that *C* is related to the concept *X* by one of the semantic relations in the defined patterns. At the same time, the concept *X* has more than one sense in the generic ontology, i.e. there is more than one semantic path originating from *X*. In this case, we compute the similarity between the new concept to be added and the concepts in the semantic paths originating from the concept *X*. Then, for the most similar semantic path(s), we update its/their hierarchy by attaching the missing concept *C* to the concept *X* using the derived semantic relation between them. Figure 3 illustrates this case. From the example discussed in section 4-C, we found that the concept "*Corporate Body*" is semantically related to the concept "*Organization*". However, the concept "*Organization*" has seven different senses in WordNet. This means that we need to find the most related sense of the concept "*Organization*" to the concept "*Corporate Body*". To do this, we compare the latter with the concepts in the semantic paths that originate from "*Organization*". Based on this comparison, the new concept "*Corporate Body*" is attached to the appropriate semantic paths as shown in Figure 3.

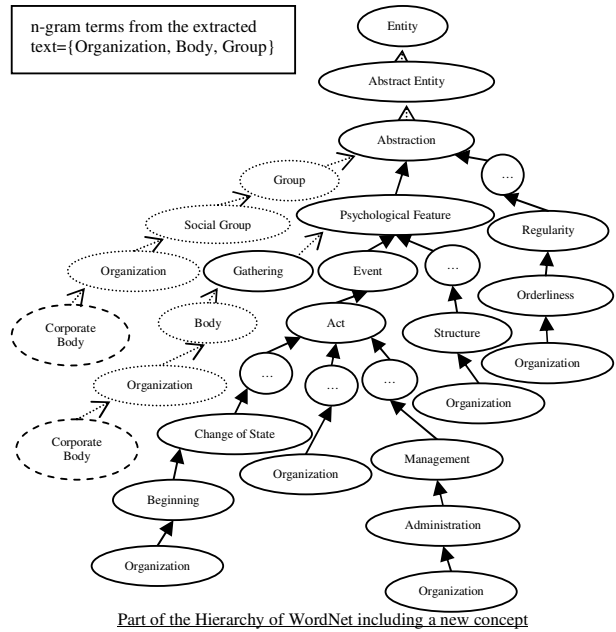

Part of the Hierarchy of WordNet including a new concept

**Fig. 3. Enriching the ontology with missing background knowledge**

**Case 3:** Concept *C* is missing from the ontology, and C is semantically related to several concepts $X_1, ..., X_n$ in the ontology. After applying the semantic relatedness measure technique, we find that *C* is related to a subset of the concepts $X_1, ..., X_i$ ($i \leq n$) by one of the semantic relations in the defined patterns. In this case, we check whether either Case 1 or Case 2 applies for each concept related to the missing concept *C*.

## 5. EXPERIMENTAL RESULTS

This section describes the experiments conducted to evaluate the performance of the statistical/semantic ontology enrichment techniques. All solutions are implemented in Java and experiments are performed on a PC with dual-core CPU (3000GHz) and 2 GB RAM. The operating system is OpenSuse 11.1. We carried out experiments using the WordNet ontology and a manually collected corpus that consists of 500 articles from different semantic domains, i.e. Animals, Science, Sports, Food kinds, Programming Languages, Countries and Cities, Universities… For each domain, text in each article is analyzed through employing GATE and the NLP techniques detailed in section 4-A. Then, we utilized both of the semantic relatedness measure and automatic pattern acquisition techniques to automatically enrich the ontology with missing background knowledge. In order to provide a ground for evaluating the quality of our results, we manually identified all concepts that are missing in the ontology. We built *expert enrichments* based on our knowledge and experience in the same fashion as highlighted in [18, 19]. Then, we compared the expert enrichments with the enrichments produced by our system. We use the precision indicator in order to measure the quality of our results. This measure is defined as follows:

**Precision (P)**: Is the percentage of the correctly enriched entities in all enriched entities. In this context, correctly enriched entities are the concepts and instances that are located in the right semantic path(s) in the ontology.

Table 3 shows the number of terms that were extracted from the articles of different domains. At this step, all of the non recognized terms are considered as missing background knowledge from the WordNet ontology. We notice that the number of terms that are recognized by GATE, the NLP techniques and the ontology (WordNet) is much smaller than the number of terms that were not recognized. This is due to several reasons. First, GATE and WordNet have limited domain coverage. Second, when utilizing the n-gram tokenization and NLP techniques, most of the bigram and trigram terms are either meaningless terms or not covered by the ontology. Therefore, we try to filter the set of non recognized terms by eliminating the meaningless ones. To do this, we employ the NH function detailed in section 4.

**Table 3. Recognized and Non Recognized Terms in the Input Texts**

| Domain Articles | # of recognized terms | # of non recognized terms |
|---|---|---|
| Articles on Animals | 744 | 6019 |

| Domain Articles | # of recognized terms | # of non recognized terms |
|---|---|---|
| Articles on Programming Languages | 328 | 3409 |
| Articles on Countries & Cities | 488 | 4012 |
| Articles on Food Kinds | 250 | 1647 |
| Articles on Universities | 454 | 3340 |
| Articles on Science | 476 | 4153 |
| Articles on Sports | 325 | 3224 |

To evaluate this algorithm, we manually eliminated the set of non recognized meaningless terms. For example, we manually eliminated the following terms from the set of non recognized terms in the Animals domain: "animals generally diurnal", "Both plant animals", "Bears aided excellent", "areas most", "caused large", etc. In the next part of the experiments, we compute the precision of the NH algorithm. It is defined as the percentage of the correctly eliminated terms given all eliminated terms. In this context, the latter are defined as the terms that were not recognized by the term recognition techniques. Tables 4 and 5 show the precision of the proposed NH algorithm in respectively eliminating and retaining non recognized terms.

**Table 4. Precision of the NH algorithm in eliminating non-recognized terms**

| Domain Articles | # of manually eliminated terms | # of N-gram Hits – Based eliminated terms | Precision (P) |
|---|---|---|---|
| Articles on Animals | 4989 | 4221 | 0.84 |
| Articles on Programming Languages | 2532 | 2108 | 0.83 |
| Articles on Countries & Cities | 2891 | 2632 | 0.91 |
| Articles on Food Kinds | 956 | 783 | 0.81 |
| Articles on Universities | 2578 | 2305 | 0.89 |
| Articles on Science | 3015 | 2789 | 0.92 |
| Articles on Sports | 3011 | 2901 | 0.96 |

**Table 5. Precision of the NH algorithm in retaining non-recognized terms**

| Domain Articles | # of manually retained terms | # of N-gram Hits –Based retained terms | Precision (P) |
|---|---|---|---|
| Articles on Animals | 1030 | 1798 | 0.57 |
| Articles on Programming Languages | 877 | 1301 | 0.67 |
| Articles on Countries & Cities | 1121 | 1380 | 0.81 |
| Articles on Food Kinds | 691 | 864 | 0.79 |
| Articles on Universities | 762 | 1035 | 0.73 |
| Articles on Science | 1138 | 1364 | 0.83 |
| Articles on Sports | 213 | 323 | 0.65 |

As shown in Tables 4 and 5, the number of manually eliminated terms is significantly larger than that of non recognized terms. This can be explained by the fact that most of the bigrams and mainly trigrams terms have no meaning. Therefore, we manually eliminated all of such terms. Figure 4 shows the percentage of error rate of the NH algorithm in eliminating non recognized terms.

Although the NH computation has a non-zero error rate in eliminating non-recognized terms, we get better performance and accuracy when using it prior to computing the NTR function. Therefore, instead of applying the NTR function on all non recognized terms, we only apply it on a filtered set of non recognized terms suggested by the NH computation.

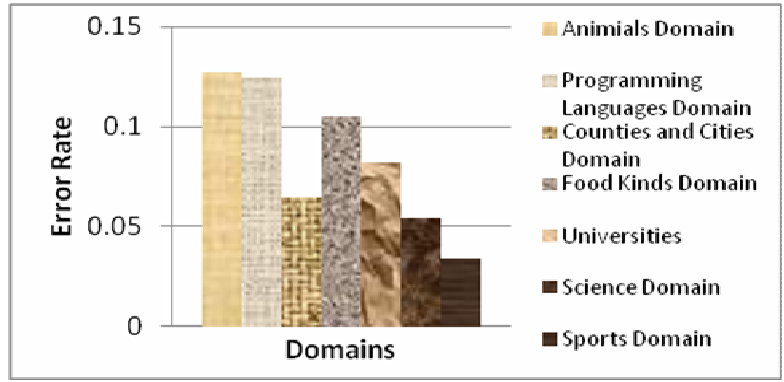

**Fig. 4. Error rate of the NH algorithm in eliminating non recognized terms**

At this phase of experiments, we are interested in the set of terms that were considered meaningful. This set is considered as the set of proposed *ontology enrichment candidates*. To enrich the ontology with the proposed set of candidates, we first employ the NTR function as detailed in section 4. In this context, semantic relatedness measures between each candidate term and other terms that are recognized by the different information extraction techniques are computed. The output of this step is a set of semantically related pairs of terms. For example, in the Sports domain, the non recognized term "Ronaldo" is suggested by the NH function to the NTR function to measure its semantic relatedness to other recognized terms. Table 6 below shows the results of computing the semantic relatedness measures between the term "*Ronaldo*" and the terms *"sport", "football", "career"* and *"Europe"*.

**Table 6. Semantic relatedness measures for the term "Ronaldo"**

| Term \ Term | *Sport* | *football* | *Career* | *Europe* |
|---|---|---|---|---|
| **Ronaldo** | 0.45 | 0.86 | 0.62 | 0.64 |

As shown in Table 6, each of the terms *"sport", "football", "career"* and *"Europe"* has a different semantic relatedness measure to the term "*Ronaldo*". We therefore utilize the automatic pattern acquisition techniques explained in section 4-C to derive the semantic relations that may hold between the pairs of semantically related terms. At this step, we notice that for some of the strongly related term pairs no results are returned. This means that none of the defined patterns could represent the semantic relation(s) that may hold between those term pairs. However, since such term pairs have very strong semantic relatedness measures, we use the proposed "Related To" relation between them and enrich the ontology with the missing term based on this relation. For example, as discussed in section 4-B, the terms "java" and "jawa" have a strong semantic relatedness measure. However, when submitting the patterns including these terms to the Google search engine, the returned results are 0 for all of the defined patterns. But, since these terms are strongly related, we enrich the ontology with the new term "jawa" using the "Related To" Relation. Figure 5 shows the result of enriching the ontology with a new term based on the "Related To" relation.

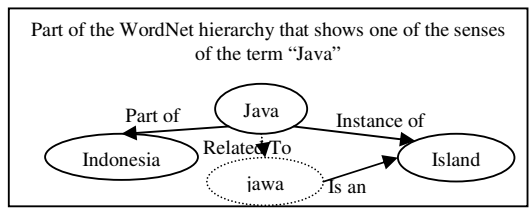

**Fig. 5. Enriching the ontology based on the "Related To" Relation**

The final empirical phase is related to evaluating the precision of the ontology enrichment process. To do this, we compare the sets of manually and automatically enriched terms for each semantic domain. The results of this step are shown in Table 7.

**Table 7. Results of Enriching the WordNet Ontology**

| Domain | Precision (%) |
|---|---|
| Articles on Animals | 81% |
| Articles on Programming Languages | 84% |
| Articles on Countries & Cities | 78% |
| Articles on Food Kinds | 69% |
| Articles on Universities | 73% |

As shown in Table 7, we were able to obtain promising precision results in general for most of the domains of interest. As we discussed earlier in section 5, those precision results are obtained based on the defined precision measure. Hence, they represent real precision measures when comparing between manually and automatically enriched entities. It is important to mention that at this phase we only considered the set of *ontology enrichment candidates* (meaningful terms obtained using the NH and NTR functions) in both sets, i.e., manually and automatically enriched entities.

Since our system showed good precision in enriching the ontology, a perspective of the work is to extend the number of used patterns. Indeed, in many cases the system suggests to enrich the ontology with a novel concept based on the "Related To" relation. For example, although the term "*Ronaldo*" has the strongest semantic relation to the term "*football*", none of the used patterns was able to represent the actual relation(s) that may hold between these two terms. And therefore, the suggestion of the system was to enrich the ontology with the term "*Ronaldo*" using the "Related To" relation. Figure 6 shows the result of this step.

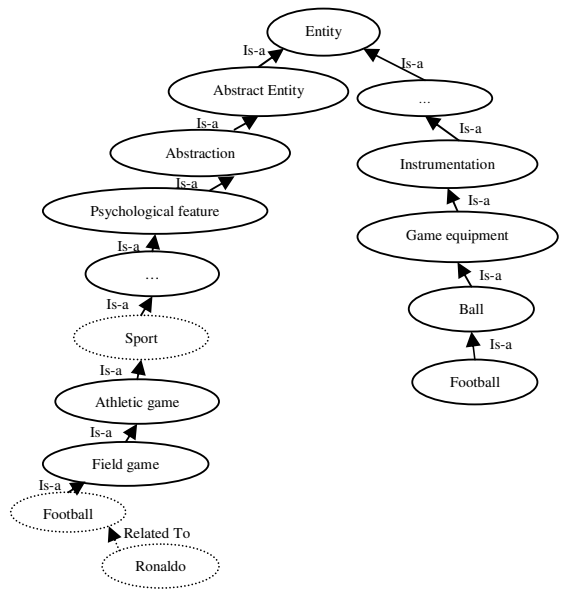

**Fig. 6. Enriching the ontology with the term "*Ronaldo*"**

As we can see from the figure, the term "*Ronaldo*" is located under the correct sense of the term "*football*". However, the "Related To" relation does not really capture the semantic relationship between these two terms. Therefore, this leads us to include both terms in the queries "Ronaldo * football" and "football * Ronaldo" then submitted to the Google search engine, where the symbol * is the wildcard operator. We will attempt to analyze the returned results in order to extract the lexical patterns between those terms.

## 6. CONCLUSION AND FUTURE WORK

In this paper, we discuss the importance of enriching existing ontology with missing background knowledge. We have surveyed state-of-the-art approaches/systems that have attempted to address this issue and classified them according to a set of categorization criteria. Additionally, we have presented our proposed solution and detailed its technical steps. The proposed solution introduces a general framework for enriching generic ontologies with missing background knowledge from the Web. In our approach, we combined semantic relatedness measures and automatic pattern acquisition techniques to extract missing entities from the WWW. In addition, multiple named entity extraction algorithms such as GATE, NLP techniques and the ontology itself are utilized. Initial experiments using 500 articles on different domains showed promising precision results. We plan to test the proposed framework using additional automatically extracted patterns to derive other types of semantic relations. We also plan to exploit other sources of knowledge such as domain specific ontologies to enrich the generic ontologies. Furthermore, we plan to do further investigation on enriching domain specific ontologies by using multiple sources of information such as other domain specific ontologies, related domain articles and multimedia information found on the WWW.